\pdfoutput=1

\documentclass[11pt]{article}

\usepackage[preprint]{acl}

\usepackage{times}
\usepackage{latexsym}
\usepackage{array}
\usepackage{tabularx}

\usepackage{xcolor}
\usepackage{cuted}        
\usepackage[most]{tcolorbox}
\usepackage{fancyvrb}
\usepackage{fvextra} 

\usepackage[T1]{fontenc}

\usepackage[utf8]{inputenc}

\usepackage{microtype}

\usepackage{inconsolata}

\usepackage{amssymb}
\usepackage{graphicx}

\usepackage{tikz}
\usepackage{pgfplots}
\usepackage{pgfplotstable}
\usepackage{makecell}
\usepackage{subfigure}
\usepackage{amsmath,amsthm,amsfonts,amssymb,bm}
\usepackage{booktabs}
\usepackage{tabularx} 
\usepackage{enumitem}
\usepackage[linesnumbered,ruled,vlined]{algorithm2e}
\usepackage{hyperref}

\usepackage{multirow}
\usepackage[normalem]{ulem}
\useunder{\uline}{\ul}{}

\usepackage{pgfplots}
\usepackage{pgfplotstable}
\usepackage{caption}
\usepackage{tikz}
\pgfplotsset{compat=1.18}

\SetKw{Return}{return}

\title{ADOPT: Adaptive Dependency-Guided Joint Prompt Optimization for Multi-Step LLM Pipelines}

\author{
Minjun Zhao \quad
Xinyu Zhang \quad
Shuai Zhang \quad
Deyang Li \quad
Ruifeng Shi \\
Huawei Poisson Lab \\
\texttt{\{zhaominjun1, zhangxinyu35, zhangshuai117, lideyang2, shiruifeng\}@huawei.com}
}

\begin{document}
\maketitle

\begin{abstract}
Multi-step LLM pipelines can solve complex tasks, but jointly optimizing prompts across steps remains challenging due to missing step-level supervision and inter-step dependency. We propose ADOPT, an adaptive dependency-guided joint prompt optimization framework for multi-step LLM pipelines. ADOPT analyzes the dependency between each LLM step and the final output, constructs a global textual gradient from final-task errors, and decomposes it into step-level local textual gradients, providing more precise optimization signals for local prompt updates. It further decouples signal estimation from prompt updating, enabling flexible integration of single-prompt optimizers, and uses a Shapley-based strategy to adaptively allocate optimization resources to high-impact steps. Experiments on real-world datasets and structurally diverse pipelines demonstrate that ADOPT is effective and robust, consistently outperforming strong prompt optimization baselines.
\end{abstract}



\section{Introduction}

\vspace{0.05in}
\noindent\textbf{Background and Motivation.}
Recent years have witnessed large language models (LLMs) achieving remarkable performance across a broad spectrum of tasks~\cite{yang2024harnessing,bubeck2023sparksartificialgeneralintelligence, achiam2023gpt, touvron2023llama}. However, despite the continuous advancement of individual LLMs, they still exhibit notable limitations when solving complex, multi-faceted problems. To address these limitations, recent research has explored multi-step LLM pipelines (i.e., workflows)~\cite{ DBLP:conf/www/LinJJHN25, yao2022react, shen2023hugginggpt, ku2025theoremexplainagent, dong2025rag, liu2025mind}, where a task is decomposed into multiple steps and executed through programs composed of coordinated LLM calls and codes. Such multi-step approaches extend the capability boundaries of LLM systems and improve robustness and success rates when tackling intricate real-world tasks~\cite{DBLP:conf/www/MaHLF25, lei2025planning, DBLP:conf/acl/HeHFLZLE25}.

\begin{figure}[t]
\centering
\includegraphics[width=0.45\textwidth]{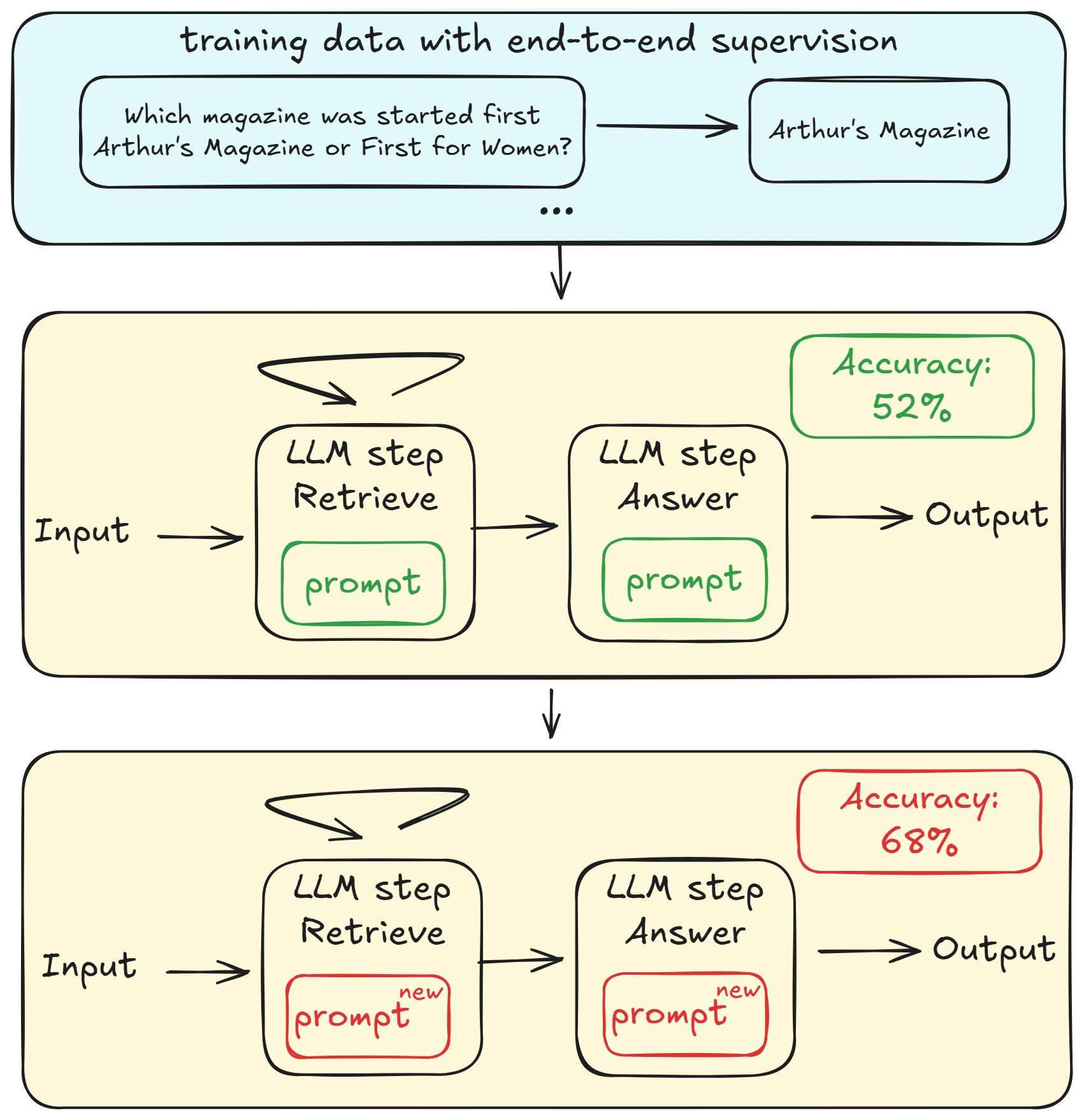}
\caption{An illustration of the prompt optimization problem in a multi-step LLM pipeline for HotPotQA.} 
\label{fig:intro}
\vspace{-4mm}
\end{figure}

Nevertheless, current multi-step LLM pipelines still face several challenges, particularly in optimizing multiple prompts. Since the prompts used at each step directly shape intermediate outputs and error propagation throughout the pipeline, effective prompt optimization is crucial for improving the end-to-end success rate of such systems. On the one hand, manual prompt optimization is highly heuristic, labor-intensive, and difficult to scale, practitioners often need to iteratively refine prompts across many steps, and modifying the prompt for one step may alter the intermediate outputs passed to later steps. Because upstream steps often cannot observe which intermediate signals are most useful for downstream components, prompts that appear well optimized in isolation can still work poorly together, so local improvements do not necessarily yield better end-to-end performance~\cite{DBLP:conf/emnlp/Opsahl-OngRPBPZ24}. On the other hand, existing automated prompt optimization methods~\cite{ProTeGi,prasad2023gripsgradientfreeeditbasedinstruction,EvoPrompt,promptbreeder} are primarily designed for optimizing prompts in single LLM calls. In multi-step LLM pipelines, only end-to-end supervision is available, and intermediate outputs are further processed before producing the final result, making it difficult to obtain labels for individual LLM calls within the pipeline.

\vspace{0.05in}
\noindent\textbf{Limitations of Existing Studies.}
Recently, several data-driven, end-to-end multi-prompt optimization methods have been proposed to address the aforementioned challenges, such as Dspy-MIPRO~\cite{DBLP:conf/emnlp/Opsahl-OngRPBPZ24}, TextGrad~\cite{DBLP:journals/nature/YuksekgonulBBLLHGZ25}, Trace~\cite{DBLP:conf/nips/ChengNS24}, and GEPA~\cite{agrawal2025gepa}. These approaches have demonstrated certain capabilities in enabling the automatic optimization of multi-step LLM pipelines. However, they also exhibit notable limitations. First, some methods~\cite{DBLP:conf/emnlp/Opsahl-OngRPBPZ24} fail to effectively utilize feedback from erroneous cases, resulting in unclear optimization directions and inefficient exploration of the prompt search space. Second, although methods using backpropagation textual gradients~\cite{DBLP:journals/nature/YuksekgonulBBLLHGZ25, DBLP:conf/nips/ChengNS24} can propagate feedback signals across steps, they rely on LLMs to perform the backward reasoning, which introduces model-induced interpretations that often distort or attenuate the gradients. Not only that, these methods struggle to determine whether an error stems from the current step or from upstream steps, leading to unstable and unreliable optimization suggestions. Besides, these methods are difficult to apply to pipelines containing looping structures, which are commonly used in complex tasks. As a result, the improvements achieved by existing end-to-end multi-prompt optimization methods are limited.

\vspace{0.05in}
\noindent\textbf{Our Solutions.}
Motivated by these limitations, we propose ADOPT, an adaptive dependency-guided joint prompt optimization framework for multi-step LLM pipelines. ADOPT consists of three components. (1) First, it analyzes execution traces to identify how each LLM step in the pipeline influences the final outcome. Based on these dependencies, ADOPT examines erroneous cases from a global perspective and distributes the high-level improvement signals to individual steps, producing a step-level adjustment direction that plays a role similar to a partial derivative with respect to each prompt. This process is analogous to computing an analytic gradient based on the functional relationship between the parameters and the final output. It avoids relying on LLMs to reason backward about where an error originates, a process that often introduces interpretation bias or severely weakens the signal. The mechanism resembles a team receiving general suggestions for improvement, where each member naturally identifies the part relevant to their own responsibilities. (2) Second, ADOPT separates the generation of these adjustment directions from the optimization procedure. Each LLM step can independently apply an appropriate single-prompt optimizer, and the system then searches for an effective combination of these local updates at the pipeline level. This design makes ADOPT flexible and compatible with a broad range of optimization strategies. (3) Third, ADOPT allocates optimization effort according to the estimated contribution of each step. Using a Shapley-based measure, LLM steps that provide greater improvement receive more optimization resources in subsequent iterations. Through dependency-guided adjustment signals, flexible local optimization, and principled resource allocation, ADOPT offers a stable and effective solution for optimizing multiple prompts in complex multi-step LLM pipelines.

\vspace{0.05in}
\noindent\textbf{Contributions.}
Our main contributions are as follows.
(1) We introduce ADOPT, an adaptive dependency-guided joint prompt optimization framework that derives step-level adjustment directions in multi-step LLM pipelines based on dependencies.
(2) We decouple adjustment-direction estimation from prompt optimization, enabling flexible use of single-prompt optimizers and effective combination of local updates.
(3) We design a Shapley-based strategy that allocates optimization resources according to each step’s contribution.
(4) Experiments on five real-world datasets and diverse pipelines show that ADOPT consistently improves end-to-end performance over existing methods.
\begin{figure*}[t]
\centering
\includegraphics[width=0.95\linewidth]{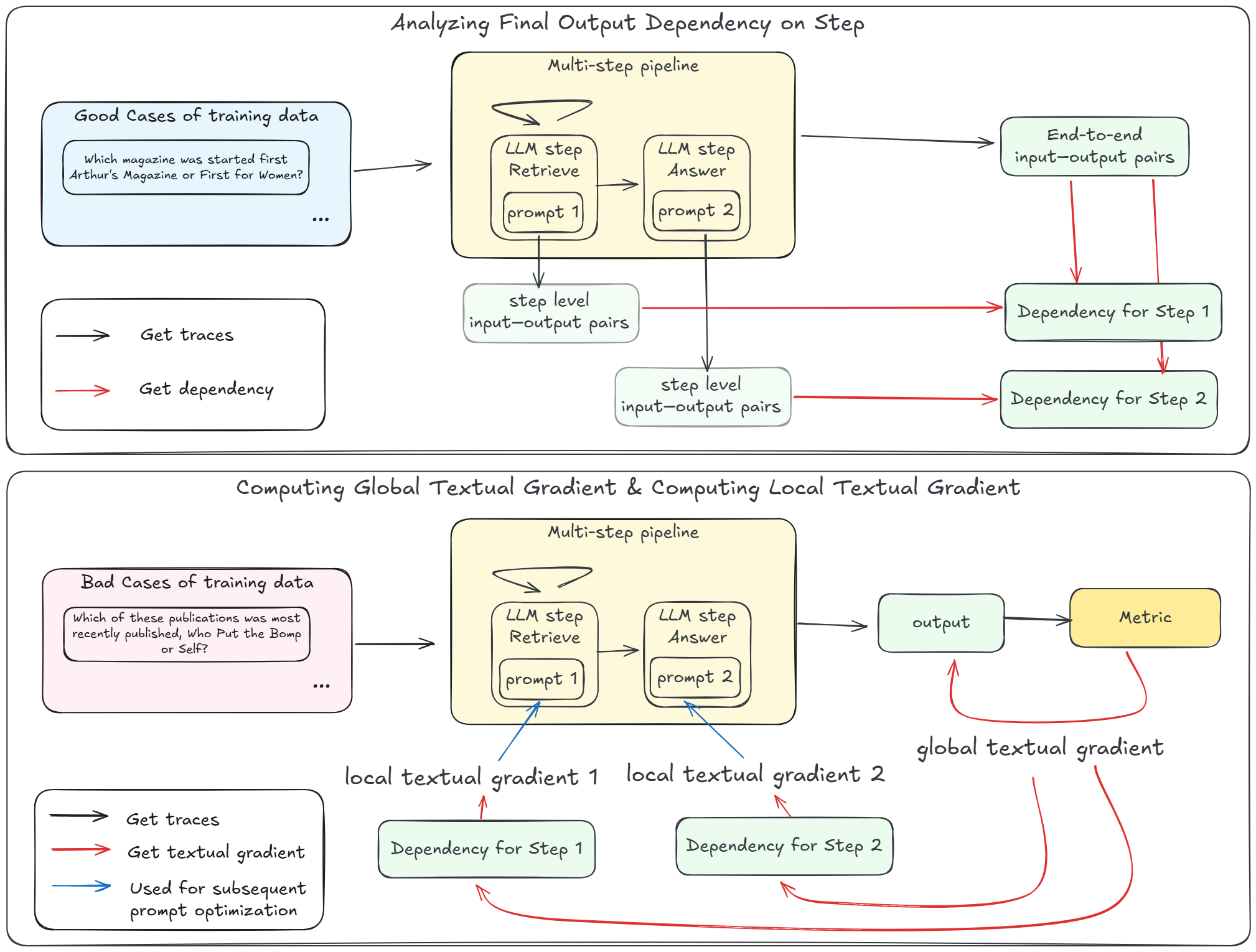}
\caption{Overview of dependency-aware textual gradient estimation in ADOPT, including final output dependency analysis, global textual gradient construction, and dependency-guided local textual gradient derivation.}
\vspace{-2mm}
\label{fig:framework}
\end{figure*}

\section{Preliminaries}

\subsection{Multi-Step LLM Pipelines}
We formalize a multi-step LLM system as
\begin{equation}
\Phi = (C, \mathcal{E}, \Pi),
\end{equation}
where $C$ denotes the non-LLM program logic of the system, including control flow, tool calls, preprocessing and postprocessing, and other operations that orchestrate the pipeline. The set $\mathcal{E}=\{\mathcal{E}_1,\dots,\mathcal{E}_m\}$ denotes the fixed LLMs used by the $m$ LLM steps, and $\Pi=\{p_1,\dots,p_m\}$ denotes the corresponding prompts.

Given an input $x$, the system executes $C$, which invokes LLM steps $(\mathcal{E}_i, p_i)$ as needed under the pipeline logic, and produces a final output $\Phi(x)$. When $C$ and $\mathcal{E}$ are fixed and only the prompts are treated as learnable parameters, we write $\Phi(x;\Pi)$.

\subsection{Problem Formulation}
The multi-step LLM system $\Phi$ takes an input $x$ and produces a final output $\Phi(x; \Pi)$, where the prompts $\Pi=\{p_1,\dots,p_m\}$ serve as the learnable parameters and all other system components remain fixed. Intermediate results generated during execution are not supervised, and only the end-to-end output is available for evaluation.

Given a training set $D_{\mathrm{train}}=\{(x_j, y_j)\}$ and a task metric $\mathcal{M}(\hat{y}, y)\in [0,1]$ that evaluates a predicted final output $\hat{y}$ against the ground-truth label $y$, where larger values indicate better performance, the goal is to optimize the prompts so that the overall system performance is maximized on the training set. Formally, we define the optimization problem as
\begin{equation}
\Pi^{*} = \arg\max_{\Pi \in S^{m}}
\sum_{(x, y)\in D_{\mathrm{train}}}
\mathcal{M}\big(\Phi(x; \Pi), y\big),
\end{equation}
where $S$ denotes the space of natural-language strings. This problem is challenging because the search space $S^{m}$ is combinatorial and extremely large, supervision is available only at the end-to-end level, and the execution of $\Phi$ is non-differentiable, making it difficult to determine which LLM steps are responsible for end-to-end errors and how their prompts should be adjusted.

\section{Methodology}
In this section, we introduce ADOPT, an adaptive dependency-guided end-to-end optimization framework for multi-step LLM pipelines. ADOPT addresses the challenge of end-to-end prompt optimization by explicitly modeling how the outputs of individual steps contribute to the final result of the pipeline and using these dependencies to guide step-wise prompt updates.

The framework operates in three stages. First, ADOPT analyzes execution traces to derive step-specific adjustment directions, which approximate analytical partial derivatives with respect to individual prompts and do not rely on backward reasoning by LLMs. Second, it decouples the estimation of these adjustment directions from the update procedure, allowing each step to employ an appropriate single-prompt optimizer while the system searches for an effective combination of local updates at the pipeline level. Third, ADOPT allocates optimization resource dynamically using Shapley value, focusing computation on the LLM steps that contribute most to performance improvement.

Together, these components enable stable, targeted, and scalable optimization of prompts in complex multi-step LLM pipelines. We use LLM-based optimizers to optimize multi-step LLM pipelines. To avoid ambiguity, we will hereafter refer to LLM-based optimizers simply as “optimizers.”

{For clarity, we provide the complete ADOPT training procedure in Appendix~\ref{sec:appendix-algorithm} (Algorithm~\ref{alg:adopt-overall}).}

\subsection{Dependency-aware Textual Gradient Estimation}

A central challenge in optimizing multi-step LLM pipelines is determining how end-to-end supervision should be decomposed into step-level update directions. ADOPT first analyzes how each step influences the final output and derives the final output dependency on every step. Based on the feedback produced by the optimizers, which describes the loss in textual form, ADOPT then generates a global optimization direction. Finally, using the inferred dependencies, ADOPT decomposes this global direction into step-level update directions.

During each training iteration, ADOPT first samples a batch of inputs from the training set and executes the pipeline $\Phi(x;\, p_1,\dots,p_m)$ on them. Based on the task metric $\mathcal{M}$, the resulting executions are categorized into good cases and bad cases, from which up to a fixed number of examples from each category are selected to form a minibatch used in the current optimization step. Figure~\ref{fig:framework} provides an overview of the dependency-aware textual gradient estimation stage.

\vspace{0.05in}
\noindent\textbf{Analyzing Final Output Dependency on Step.}
To simulate the effect of computing analytical partial derivatives, ADOPT must first determine how each LLM step functionally influences the final pipeline output. ADOPT begins with a direct analysis stage, in which an optimizer $E_1$ examines the workflow code together with the prompts of all steps to infer the overall task and the intended role of each step within the pipeline.

Building on this structural understanding, ADOPT then performs a data-driven analysis using the traces collected from good cases. For each LLM step, optimizer $E_2$ is provided with the direct-analysis results, the step’s input–output pairs, and the corresponding end-to-end traces. The optimizer analyzes how variations in the step’s output affect the final result, yielding a final output \emph{dependency} on that step. This dependency characterizes the step’s functional contribution to successful execution and its influence on the end-to-end behavior.

\vspace{0.05in}
\noindent\textbf{Computing Global Textual Gradient.}
For each bad case, ADOPT first uses optimizer $E_3$ to construct a textual discrepancy $\mathcal{L}$ between the pipeline output and the ground truth, guided by the task metric $\mathcal{M}$. This discrepancy, referred to as \emph{textual loss} or \emph{feedback} in our setting, explicitly enumerates the errors responsible for the score reduction and provides a fine-grained and interpretable description of the mismatch, serving as the natural-language counterpart of an error signal.

ADOPT then invokes optimizer $E_4$ to transform this textual loss into a \emph{global textual gradient}. $E_4$ performs a diagnostic analysis that explains why each discrepancy arises, producing a global description of the failure mode of the final output. This yields a distilled correction direction expressed at the level of the final output itself. The resulting global textual gradient functions as an analogue of the derivative of $\mathcal{L}$ with respect to the textual loss, that is,
$g^{\text{global}} = \nabla_{\text{Output}}^{\text{text}} \mathcal{L}.$

\vspace{0.05in}
\noindent\textbf{Computing Local Textual Gradient.}
Given the global textual gradient and the final output dependencies inferred for each step, ADOPT uses optimizer $E_5$ to compute a \emph{local textual gradient} for every node. For each step, $E_5$ receives the global textual gradient, the step’s final output dependency, the step-level input–output pair, and the final output, and then produces a step-specific natural-language optimization direction that constitutes the local textual gradient. Formally, this direction serves as the textual analogue of an analytical partial derivative,
$g^{\text{local}}_i = \nabla_{p_i}^{\text{text}} \mathcal{L}.$

\subsection{Prompt Optimization for Pipeline}
After deriving step-level local textual gradients, the remaining challenge is to convert these textual optimization signals into executable prompt updates while achieving effective coordination across a multi-step LLM pipeline. Pipeline-level prompt optimization requires not only that each step independently updates its prompt according to its local signal, but also that the resulting prompts interact coherently in an end-to-end manner. To address this, ADOPT adopts a prompt optimization strategy that combines step-level prompt optimization guided by local textual gradients with pipeline-level prompt selection based on search algorithms.

\vspace{0.05in}
\noindent\textbf{Decoupled Step-level Prompt Optimization.}
For each bad case, ADOPT uses the local textual gradient of step $i$ to generate a \emph{revised step output} that represents the desired behavior of this step under the given input. This revised output specifies how the step’s output should change in order to reduce the end-to-end loss and correct the final pipeline error. Formally, for step $i$ with input $x_i$, the revised output $\hat{y}_i$ is generated by optimizer $E_6$ guided by the local textual gradient.

Across multiple bad cases, ADOPT aggregates pairs of step-level inputs and their corresponding revised step outputs, forming a step-specific dataset
\begin{equation}
\mathcal{D}_i = \{(x_i^{(k)}, \hat{y}_i^{(k)})\}_{k=1}^{K}.
\end{equation}
This dataset serves as supervision for updating the prompt $p_i$ of LLM step $i$ using a single-prompt optimization procedure.

Crucially, ADOPT decouples the generation of optimization directions from the choice of the prompt optimization algorithm. For each LLM step, any existing single-prompt optimizer can be applied to $\mathcal{D}_i$ to update $p_i$, without modifying the dependency analysis or textual gradient computation stages. This design makes the framework modular and extensible, allowing different prompt optimizers to be incorporated as interchangeable components.

In this work, we consider two representative optimizers. The first is an instruction optimizer that updates only the instruction component of the prompt. The second jointly optimizes the instruction and automatically selects representative examples to be included as in-context learning demonstrations. Each round of step-level optimization produces multiple candidate prompts for each LLM step, which are subsequently evaluated and coordinated at the pipeline level.

\vspace{0.05in}
\noindent\textbf{Global Prompt Selection.}
The step-level prompt optimization procedure produces multiple candidate prompts for each LLM step, which gives rise to a pipeline-level selection problem. Let $\mathcal{P}_i = \{p_i^{(1)}, \dots, p_i^{(n_i)}\}$ denote the candidate prompt set generated for step $i$. The objective is to select a prompt configuration $(p_1, \dots, p_m) \in \mathcal{P}_1 \times \cdots \times \mathcal{P}_m$ that maximizes the end-to-end performance of the pipeline. Since the effect of a prompt at one step may depend on the prompts chosen at other steps, this problem cannot be decomposed into independent per-step decisions and instead constitutes a combinatorial optimization problem over the joint prompt space.

We formulate pipeline-level prompt selection as a search problem and apply Bayesian Optimization~\cite{snoek2012practical} to efficiently explore prompt configurations under a limited budget.

\subsection{Shapley-based Resource Allocation}

Prompt updates in multi-step LLM pipelines typically yield uneven returns across steps. After several optimization rounds, certain steps may become saturated and contribute little additional improvement, while other steps remain critical to the end-to-end metric. Treating all steps equally, therefore, wastes optimization budget. ADOPT addresses this issue by allocating step-level optimization resources according to each step’s estimated contribution to end-to-end performance, and adopts Shapley-style attribution as a principled measure of contribution.

Before optimization begins, each step has the same optimization resources, i.e., how many candidates can be generated during step-level optimization. 
At the end of each round, the global prompt selector evaluates a set of prompt configurations and records their end-to-end scores under the task metric $\mathcal{M}$. For each step $i$, we consider two prompt states, a \emph{weak} prompt $p_i^{\mathrm{weak}}$ representing the prompt used in the previous round, and a \emph{strong} prompt $p_i^{\mathrm{strong}}$ representing the best-performing prompt selected from the current candidate set $\mathcal{P}_i$. 
Among all evaluated configurations, ADOPT considers only those in which each step adopts either its weak or strong prompt. These configurations can thus be represented by a coalition $S \subseteq \{1,\dots,m\}$, where $i \in S$ indicates that step $i$ uses $p_i^{\mathrm{strong}}$ and $i \notin S$ uses $p_i^{\mathrm{weak}}$. Let $P(S)$ denote the corresponding prompt assignment, and define the value function
\begin{equation}
v(S) = \mathcal{M}\big(\Phi(x; P(S))\big).
\end{equation}
The Shapley value for step $i$ measures its average marginal contribution across all coalitions,
\begin{equation}
\phi_i
= \sum_{S \subseteq [m]\setminus\{i\}}
w(S)\,
\big(v(S \cup \{i\}) - v(S)\big),
\end{equation}
where $w(S) = \frac{|S|!(m-|S|-1)!}{m!}$. Computing $\{\phi_i\}$ exactly requires evaluating $v(S)$ for an exponential number of coalitions, which is prohibitively expensive because each evaluation corresponds to running the full pipeline. ADOPT therefore estimates step contributions using Kernel SHAP~\cite{lundberg2017unified}, which approximates Shapley values from a limited number of observed coalitions. Concretely, ADOPT applies Kernel SHAP to the prompt configurations that have already been evaluated during the pipeline-level search. Each evaluated configuration provides a binary indicator vector $z \in \{0,1\}^m$ for its coalition membership and an observed score $v(z)$. Kernel SHAP then fits a weighted linear model over these samples and uses the fitted coefficients as efficient approximations of Shapley-style contributions.

The resulting contribution estimates are used only for resource allocation in subsequent rounds. Steps with larger estimated contributions are assigned larger candidate budgets, while steps with smaller contributions receive fewer candidates. This reallocation preserves the overall evaluation budget. The number of pipeline-level prompt configurations evaluated per round is kept fixed, and no additional pipeline executions are introduced for contribution estimation. In this way, ADOPT concentrates optimization effort on the steps that most influence end-to-end performance while maintaining stable computational cost across rounds.

We further report token and time consumption in Appendix~\ref{sec:Efficiency}, showing that adaptive allocation improves the cost-effectiveness of optimization.
\section{Experiments}

\begin{table}[t]
\centering
\small
\caption{Information of five diverse multi-step pipelines on real-world datasets.}
\label{tab:table-data}
\begin{tabular}{|c|c|c|c|}
\hline
\textbf{Benchmark} & \textbf{Domain} & \textbf{Steps} & \textbf{Loop} \\ \hline
Amazon    & Recommendation   & 3     & None     \\ \hline
HotPotQA    & \begin{tabular}[c]{@{}c@{}}Multi-hop\\ Question Answering\end{tabular}   & 2    & Exist    \\ \hline
HoVer      & Fact Verification    & 5     & None    \\ \hline
Emergency  & Digital Assistance   & 4    & Exist   \\ \hline
Harmful & \begin{tabular}[c]{@{}c@{}}Social Abuse\\ Detection\end{tabular} & 4  & Exist  \\ \hline
\end{tabular}
\end{table}

\begin{table*}[]
\centering
\caption{Overall performance of ADOPT and baselines on different real-world datasets.}
\label{tab:Overall}
\vspace{-2mm}
\begin{tabular}{lccccccc}
\toprule
Dataset & No-COT & COT & TextGrad &	Trace & MIPRO & GEPA & ADOPT\\
\midrule
Amazon	& 0.88	& 0.89	& 0.90	& 0.92	& 0.91	& 0.92	& \textbf{0.94}	\\
HotPotQA & 0.52 & 0.58 & / & / & 0.62 & 0.63 & \textbf{0.69} \\
HoVer  & 0.55 & 0.57 &  0.58 & 0.59 & 0.62 & 0.65 & \textbf{0.71}  \\
Emergency	& 0.71	& 0.73	& /	& /	& 0.78	& 0.83	& \textbf{0.89}	\\
Harmful	& 0.58	& 0.61	& /	& /	& 0.67	& 0.71	& \textbf{0.75}	\\
\bottomrule
\end{tabular}
\end{table*}

\subsection{Experimental Setup}
We compare ADOPT with six methods: Chain-of-Thought (CoT)~\cite{DBLP:conf/nips/Wei0SBIXCLZ22}, no COT, Trace~\cite{DBLP:conf/nips/ChengNS24}, TextGrad~\cite{yuksekgonul2024textgrad}, MIPRO~\cite{DBLP:conf/emnlp/Opsahl-OngRPBPZ24}, GEPA~\cite{agrawal2025gepa}. For CoT, we add "let's think step by step" to the prompt of each agent in the tested programs.

We implement five multi-step pipelines to thoroughly evaluate the effectiveness of our proposed ADOPT and baselines. And we use five real-world datasets, i.e., Amazon~\cite{DBLP:journals/corr/abs-2403-03952}, HotPotQA~\cite{yang2018hotpotqa}, HoVer~\cite{jiang2020hover}, Emergency~\cite{zhu2025heal}, and Harmful~\cite{DBLP:conf/cvpr/KhannaRCYGCKCBM24} as benchmarks for different multi-step pipelines separately. We show the information of our diverse multi-step pipelines in Table~\ref{tab:table-data}, and we show the implementation details in the Appendix~\ref{sec:appendix}. Since different programs run on different datasets and measure the performance on different values, we introduce each metric used in the experiments. For Amazon dataset, the metric is $1-MAE$~\cite{DBLP:conf/naacl/WangJCYZCFLHY24}, where MAE is normalized Mean Absolute Error of rating. For other datasets, we use accuracy as the metric.


In the main experiments, we use Qwen2.5-72B-Instruct~\cite{team2024qwen2} as the default model configuration. For all LLM calls, the temperature is set to 0 and top-$p$ is set to 1.
Each training iteration executes the pipeline on a batch of 50 cases sampled from the training set, from which up to 5 good cases and 5 bad cases are subsampled to form the minibatch used for prompt optimization. All reported metrics are evaluated on the test set. Unless otherwise specified, all experiments are repeated three times with different random seeds, and the reported results are averaged over these runs.

We also analyze optimization efficiency, cross-model robustness, and a step-level case study in the appendix, which provide further evidence for the proposed dependency-guided optimization mechanism.

\subsection{Main Results}
Table~\ref{tab:Overall} reports the overall performance of ADOPT and baseline methods on two real-world multi-step benchmarks. We do not report results for TextGrad and Trace on HotPotQA, Emergency, and Harmful datasets because these methods are designed primarily for LLM workflows with directed acyclic graph (DAG) structures and do not naturally support pipelines with looping or iterative control flow. 

\begin{figure}[t]
    \centering
    \subfigure[HotPotQA]{
        \includegraphics[width=0.46\linewidth]{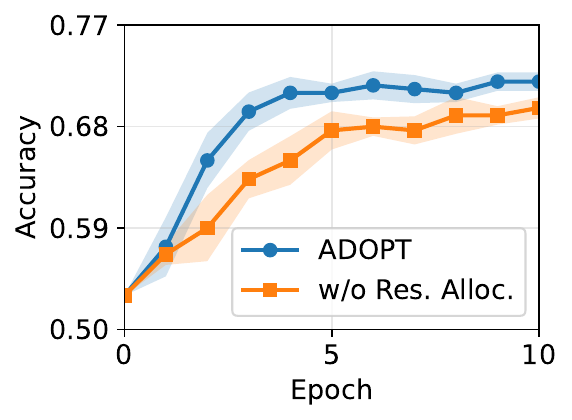}
    }
    \hfill
    \subfigure[HoVer]{
        \includegraphics[width=0.46\linewidth]{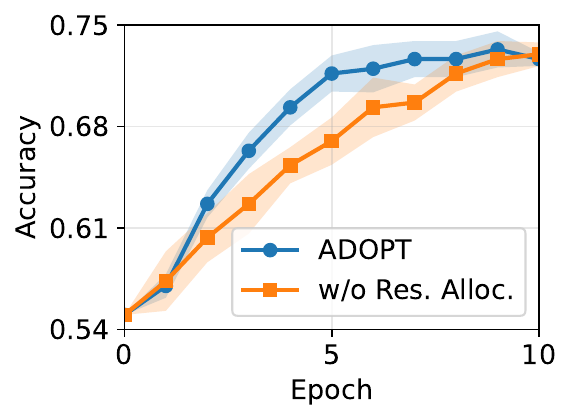}
    }
    \caption{Convergence analysis of ADOPT on HotPotQA and HoVer.}
\label{fig:convergence}
\end{figure}

Across all evaluated datasets, ADOPT consistently achieves the strongest end-to-end performance, demonstrating its effectiveness for optimizing multi-step LLM pipelines under end-to-end supervision. Compared with prompting-based baselines such as CoT, ADOPT yields substantial performance gains, indicating that simply encouraging step-by-step reasoning is insufficient to address errors arising from complex interactions among multiple pipeline steps.
ADOPT also outperforms existing multi-prompt optimization methods, including TextGrad, Trace, MIPRO, and GEPA. TextGrad and Trace rely on LLMs to perform backward reasoning in order to generate step-level optimization signals, which introduces model-induced interpretations that can distort or attenuate the feedback after several steps. Moreover, these methods often struggle to disentangle whether an error originates from a specific step or from upstream dependencies, leading to unstable or inconsistent optimization behavior. MIPRO generates candidate prompts only from correctly solved examples and selects an optimal combination among them, without iteratively refining prompts based on bad cases. GEPA incorporates feedback from bad cases, but directly optimizes prompts using only final-output feedback, analogous to the textual loss in our framework, without explicitly modeling how individual steps should be adjusted. In contrast, ADOPT leverages textual loss to derive global textual gradient and uses dependencies to decompose end-to-end supervision into step-specific optimization signals, i.e., local textual gradients, enabling more precise and coordinated step-level optimization across the pipeline, which leads to consistently better end-to-end performance. Taken together, these results indicate that ADOPT provides a more effective mechanism for improving end-to-end performance in multi-step LLM systems than existing methods.

\subsection{Convergence Analysis}
To examine the convergence behavior of ADOPT, we use 300 validation cases from HotPotQA and HoVer, respectively, and track the end-to-end accuracy over successive optimization iterations. As shown in Figure~\ref{fig:convergence}, ADOPT improves steadily from the initial iterations, with accuracy increasing rapidly in the early stage and continuing to improve in later iterations. Although minor fluctuations appear in the later stage, the overall trend remains stable, and the performance does not degrade as optimization proceeds. This suggests that the step-level optimization signals derived by ADOPT are sufficiently informative to support effective iterative updates in multi-step pipelines. In addition, ADOPT with resource allocation converges faster than the variant without it, indicating that contribution-aware budget allocation helps improve optimization efficiency.

\subsection{Ablation Studies}

\begin{table}[tp]
\centering
\caption{Comparison of different step-level prompt optimizers on two datasets.}
\begin{tabular}{lcc}
\toprule
Dataset	& ADOPT-Instruct	& ADOPT-Joint \\
\midrule
HotPotQA	& 0.67	& \textbf{0.69} \\
HoVer	& 0.69	& \textbf{0.71} \\
\bottomrule
\end{tabular}
\label{tab:prompt-optimizer}
\end{table}
\vspace{0.05in}

\noindent\textbf{Effect of Different Step-level Prompt Optimizers.}
ADOPT decouples dependency-aware textual gradient estimation from the choice of the step-level prompt optimization algorithm, allowing different single-prompt optimizers to be applied without modifying the upstream analysis. To assess the effect of this design, we instantiate ADOPT with two representative optimizers at the step level: an instruction-only optimizer (i.e., ADOPT-Instruct) that updates the natural-language instruction in each prompt, and a joint instruction-and-example optimizer (i.e., ADOPT-Joint) that additionally selects representative in-context learning examples.

The results are shown in Table~\ref{tab:prompt-optimizer}. From the table, both optimizers consistently improve end-to-end performance over the baselines, while the joint optimizer achieves further gains, indicating that enriching step prompts with selected examples can better capture step-specific behaviors when accurate local supervision is available. These results demonstrate that ADOPT’s performance is not tied to a particular prompt optimizer and that its decoupled formulation enables the framework to benefit from more expressive optimization strategies as they become available.

\begin{table}[t]
\centering
\caption{Fine-grained ablation of ADOPT. All variants are run for 10 optimization iterations, and we report the best accuracy achieved across iterations.}
\vspace{-2mm}
\label{tab:component-ablation}
\small
\setlength{\tabcolsep}{6pt}
\begin{tabular}{lcc}
\toprule
Method & HotPotQA & HoVer \\
\midrule
ADOPT & \textbf{0.69} & \textbf{0.71} \\
w/o Dependency-guided Est. & 0.64 & 0.67 \\
w/o Global Prompt Selection & 0.62 & 0.66 \\
w/o Resource Allocation & 0.68 & \textbf{0.71} \\
\bottomrule
\end{tabular}
\vspace{-2mm}
\end{table}

\noindent\textbf{Fine-grained Component Ablation.}
Table~\ref{tab:component-ablation} shows a fine-grained ablation of ADOPT on HotPotQA and HoVer. We study three key components: dependency-guided estimation, global prompt selection, and resource allocation. All variants are run for 10 optimization iterations, and we report the best accuracy across iterations.

Removing dependency-guided estimation degrades performance on both datasets, dropping from 0.69 to 0.64 on HotPotQA and from 0.71 to 0.67 on HoVer. This shows that dependency-aware decomposition provides more effective step-level optimization signals.

Removing global prompt selection further reduces performance to 0.62 on HotPotQA and 0.66 on HoVer. This is because candidate generation is stochastic, and locally improved prompts do not always yield the best pipeline-level combination. In some cases, keeping the current prompt is better than updating it with a newly optimized candidate.

In contrast, removing resource allocation has little impact on the final best accuracy, with results of 0.68 on HotPotQA and 0.71 on HoVer. This suggests that resource allocation mainly improves optimization efficiency rather than final accuracy.

\section{Related Work}
\vspace{0.05in}
\noindent\textbf{Prompt Optimization for Single-Step Tasks.}
Prompt optimization for single-step tasks typically refines a static prompt for a single LLM call. Representative methods include search-based approaches such as APE~\cite{APE} and OPRO~\cite{OPRO}, evolutionary methods such as EvoPrompt~\cite{EvoPrompt}, and gradient-inspired approaches such as ProTeGi~\cite{ProTeGi}. AMPO~\cite{AMPO} further improves robustness by exploring multiple optimization trajectories in parallel. While effective for single-step settings, these methods do not directly address prompt optimization in multi-step LLM pipelines with inter-step dependencies.

\vspace{0.05in}
\noindent\textbf{Prompt Optimization in Multi-Step Pipelines.}
Recent work has extended prompt optimization to multi-step LLM workflows. TextGrad~\cite{yuksekgonul2024textgrad} and Trace~\cite{DBLP:conf/nips/ChengNS24} assume structured workflows and propagate feedback backward to produce step-level update signals. MIPRO~\cite{DBLP:conf/emnlp/Opsahl-OngRPBPZ24} optimizes module-level instructions and demonstrations by selecting prompt configurations from candidate sets. GEPA~\cite{agrawal2025gepa} leverages erroneous cases and final-output feedback to refine prompts. In contrast, ADOPT explicitly decomposes end-to-end supervision into dependency-aware, step-specific optimization directions, enabling more reliable prompt updates in multi-step pipelines.

\section{Conclusion}

We propose ADOPT, an adaptive dependency-guided joint prompt optimization framework for multi-step LLM pipelines. ADOPT decomposes end-task supervision into step-level optimization signals through dependency analysis, supports flexible integration of existing single-prompt optimizers, and improves optimization efficiency with Shapley-based resource allocation. Experiments show that ADOPT consistently improves end-to-end performance.
\section*{Limitations}
Although ADOPT demonstrates strong performance in optimizing multi-step pipelines, it inevitably incurs a relatively high number of LLM calls due to its iterative optimization process. To mitigate this, we design a Shapley value-based resource allocation module that prioritizes optimization efforts toward the most impactful LLM step. This allows ADOPT to reduce redundant computations and achieve the same level of performance with fewer LLM calls compared to uniform optimization strategies. Despite these improvements, the overall resource consumption remains non-trivial, especially for large-scale or real-time applications.
Reducing this cost further, for example through more efficient search and resource allocation strategies, remains an important direction for future work.

\bibliography{custom}

\appendix

\newpage
\onecolumn
{
\section{Overall ADOPT Procedure}
\label{sec:appendix-algorithm}
Algorithm~\ref{alg:adopt-overall} presents the overall procedure of ADOPT. At each optimization round, ADOPT runs the current pipeline on a minibatch, collects execution traces, and separates the cases into good and bad ones according to the end-to-end metric. For bad cases, it estimates final-output dependencies, constructs a global textual gradient from end-to-end errors, and decomposes it into local textual gradients for individual steps. These local gradients are used to build step-level supervision, which enables each step prompt to be updated by a single-prompt optimizer under an adaptive budget. ADOPT then performs pipeline-level search to select the best prompt combination and applies Kernel SHAP to estimate step contributions, which are further used to reallocate optimization resources in the next round. 

\begin{algorithm}[t]
\caption{ADOPT: Adaptive Dependency-Guided Joint Prompt Optimization for Multi-Step LLM Pipelines}
\label{alg:adopt-overall}
\DontPrintSemicolon
\KwIn{Pipeline $\Phi=(C,M,\Pi)$, training set $D_{\mathrm{train}}$, metric $\mathcal{M}$, rounds $R$, per-round budget $B$}
\KwOut{Optimized prompts $\Pi^*$}

Initialize prompts $\Pi^{(0)}$, candidate allocation $\{b_i^{(0)}\}_{i=1}^{m}$ with $\sum_i b_i^{(0)}=B$\;
\For{$r=1$ \KwTo $R$}{
Sample minibatch and execute $\Phi(x;\Pi^{(r-1)})$ to collect traces\;
Partition traces into good/bad cases by metric $\mathcal{M}$\;
Estimate final-output dependencies for each bad case\;
Construct global textual gradients from end-to-end errors\;
Decompose global gradients into local textual gradients per step\;
Generate revised step outputs and build step datasets $\{\mathcal{D}_i\}$\;
Optimize each step prompt with single-prompt optimizer under budget $b_i^{(r-1)}$ to get candidate sets $\{\mathcal{P}_i\}$\;
Run pipeline-level search over combinations in $\mathcal{P}_1\times\cdots\times\mathcal{P}_m$ and select best prompt set $\Pi^{(r)}$\;
Estimate step contributions with Kernel SHAP from evaluated combinations\;
Update allocation $\{b_i^{(r)}\}$ for next round based on contribution estimates\;
}
\Return{$\Pi^{(R)}$}\;
\end{algorithm}

\newpage
\onecolumn

\begin{table*}[t]
\centering
\caption{Token consumption and optimization time of different methods on HotPotQA and HoVer, using parallelism of 10. Token counts are reported in millions (M).}
\vspace{-2mm}
\label{tab:efficiency-overhead}
\begin{tabular}{llcccc}
\toprule
Dataset & Method & Input (M) & Output (M) & Total (M) & Opt. Time (h) \\
\midrule
\multirow{3}{*}{HotPotQA}
& MIPRO & 11.21 & \textbf{0.57} & 11.79 & 2.32 \\
& GEPA  & \textbf{3.98}  & 0.67 & \textbf{4.65}  & 3.88 \\
& ADOPT & 4.86  & 0.66 & 5.52  & \textbf{2.20} \\
& \quad w/o Resource Allocation & 7.42 & 1.13 & 8.56 & 3.82 \\
\midrule
\multirow{3}{*}{HoVer}
& MIPRO & 71.71 & \textbf{3.82} & 75.53 & 15.34 \\
& GEPA  & \textbf{27.13} & 3.99 & \textbf{31.11} & 25.67 \\
& ADOPT & 36.92 & 6.38 & 43.31 & \textbf{10.89} \\
& \quad w/o Resource Allocation & 52.11 & 10.21 & 62.32 & 16.97 \\
\bottomrule
\end{tabular}%
\end{table*}

\section{Efficiency and Overhead Analysis}
\label{sec:Efficiency}
To address concerns about practical overhead, we compare ADOPT with MIPRO and GEPA on HotPotQA and HoVer using total token consumption and end-to-end optimization wall-clock time. All methods use the same backbone model and the same parallelism level (10 threads). We also include ADOPT w/o Resource Allocation to evaluate the impact of the Shapley-based resource allocation module on optimization efficiency.

As shown in Table~\ref{tab:efficiency-overhead}, ADOPT achieves a favorable efficiency and effectiveness trade-off. Compared with GEPA, it consistently reduces optimization time while maintaining comparable token consumption. Compared with MIPRO, it requires much fewer tokens and is also faster on both datasets. In addition, removing the resource allocation module increases both token consumption and optimization time, showing that contribution-aware resource allocation further reduces practical optimization overhead. We attribute this advantage to the design of ADOPT. The dependency-aware optimization produces more accurate step-level update signals, which reduces ineffective prompt exploration, and the resource allocation module allocates more budget to the steps that have larger impact on final performance.

\newpage
\onecolumn

\section{Case Study}
\label{sec:appendix-case-study}


We present a case study on HotPotQA to illustrate how ADOPT optimizes a \emph{non-final} LLM step using only end-to-end supervision. Specifically, we examine the first LLM step, \texttt{generate\_query}, in one optimization iteration. This step does not directly produce the final answer; instead, it generates retrieval queries that influence downstream evidence collection and ultimately affect answer quality. Therefore, this example is particularly suitable for demonstrating the central challenge addressed by ADOPT: \emph{how to optimize an intermediate LLM node when only the final workflow output is supervised}.

As described in Appendix~\ref{sec:appendix}, the HotPotQA pipeline contains two LLM steps: \texttt{generate\_query} and \texttt{generate\_answer}. The former iteratively generates retrieval queries conditioned on the current context and the question, while the latter produces the final answer based on the accumulated retrieved context. In this case study, the supervision signal comes only from the mismatch between the final answer and the gold answer; no gold query or step-level target is available for \texttt{generate\_query}. ADOPT addresses this challenge by deriving dependency-aware local optimization signals from the final error.

We first show the original prompt of the \texttt{generate\_query} step before optimization.

\begin{tcolorbox}[breakable,colback=gray!10!white, colframe=gray!80!black,
  title=Prompt of \texttt{generate\_query} Before Optimization, boxrule=0.5pt, arc=2mm, left=2mm, right=2mm, top=1mm, bottom=1mm]
\begin{Verbatim}[fontsize=\small, breaklines=true]
Your input fields are:
1. `context` (str): 
2. `question` (str):
Your output fields are:
1. `reasoning` (str): 
2. `search_query` (str):
All interactions will be structured in the following way, with the appropriate values filled in.

[[ ## context ## ]]
{context}

[[ ## question ## ]]
{question}

[[ ## reasoning ## ]]
{reasoning}

[[ ## search_query ## ]]
{search_query}

[[ ## completed ## ]]
In adhering to this structure, your objective is: 
        Given the fields `context`, `question`, produce the fields `search_query`
\end{Verbatim}
\end{tcolorbox}

To optimize this step without direct supervision, ADOPT first analyzes how the output of \texttt{generate\_query} functionally contributes to the final answer. This dependency analysis is inferred from good execution traces and serves as the basis for decomposing end-to-end feedback into step-specific update directions.

\begin{tcolorbox}[breakable,colback=gray!10!white, colframe=gray!80!black,
  title=Dependency Analysis for \texttt{generate\_query}, boxrule=0.5pt, arc=2mm, left=2mm, right=2mm, top=1mm, bottom=1mm]
\begin{Verbatim}[fontsize=\small, breaklines=true]
### Analysis Summary

#### 1. Based on the good cases, what specific duty does this LLM node perform in the workflow?
The primary responsibility of this LLM node is to generate a new query (`search_query`)
based on the existing retrieved content (`context`) and the user’s question (`question`),
so that more relevant information can be retrieved in subsequent steps.

Specifically:
* Input: `context` and `question`
* Output: `search_query`

The contribution of this node’s output to the final result includes:
* Supplementary Information: The generated `search_query` can help retrieve more
  information relevant to the user’s question, enriching the final answer.
* Refined Search: By analyzing the existing context and the user’s question, it
  generates a more precise query, improving the accuracy of subsequent retrievals.

#### 2. Based on the good cases, what is the correlation between this LLM node’s output and the final output?
* Information Supplementation: In multiple cases, the generated `search_query`
  helps retrieve key information that plays a decisive role in the final answer.
  For example, in the “Mel McDaniel’s album” case, the generated query helped
  locate information about Don Williams.

* Precise Matching: The generated query usually aligns more precisely with the
  user’s question, enabling subsequent retrievals to find more relevant contextual
  content. For example, in the “first UFC match in history” case, the query helped
  locate information about the American fighter.

#### 3. Based on the good cases, how might changes in the LLM node’s output positively or negatively affect the final workflow output?
* Positive Impact:
  * More precise queries can better match the user’s question, resulting in
    more relevant retrieved content and higher-quality final answers.
  * Supplementing key information can improve both the accuracy and completeness
    of the final response.

* Negative Impact:
  * Imprecise queries may retrieve unrelated or less relevant content, reducing
    the quality of the final answer.
  * Failure to cover all critical information points may cause important details
    to be missing, reducing completeness and accuracy.

#### Summary
This node generates new queries that supplement and refine subsequent retrieval content.
Its output has a significant effect on the final result: the more precise the generated
query, the higher the quality of the final answer.
\end{Verbatim}
\end{tcolorbox}

After establishing the dependency of the final output on this step, ADOPT examines a bad case through the discrepancy between the workflow output and the ground-truth answer. This discrepancy is the textual loss, which remains defined only at the final-output level.

\begin{tcolorbox}[breakable,colback=gray!10!white, colframe=gray!80!black,
  title=Textual Loss from Final Output, boxrule=0.5pt, arc=2mm, left=2mm, right=2mm, top=1mm, bottom=1mm]
\begin{Verbatim}[fontsize=\small, breaklines=true]
1. The key information in the current answer is “The game...launched on iOS and Maemo devices,”
   where “iOS and Maemo devices” refers to operating systems or device platforms, whereas the
   question asks “Where,” which should refer to a geographic region. Compared with the gold
   answer “New Zealand,” the current output confuses “release platform” with “initial release
   country,” causing a semantic deviation.

2. The answer does not mention any geographic location, especially the core entity “New Zealand.”
   This omission directly makes the answer factually incorrect and severely harms the score.

3. The answer focuses on technical release details (i.e., the types of devices on which the game
   ran), rather than the geographic release scope targeted by the question. This suggests that the
   model misunderstood the intent of the “Where” question and needs to shift its interpretation
   from “platform” to “country.”

4. Although the output format is structurally correct as a dictionary {'answer': '...'}, the content
   is entirely inconsistent with the correct fact and thus constitutes a substantive error. The retrieval
   stage must obtain information about the “initial release country,” rather than remain limited to
   device compatibility.

5. In the multi-step retrieval process, the query strategy for “Where” questions should explicitly
   include geographically oriented keywords such as “country,” “first released in,” and
   “launch location,” so as to increase the probability of retrieving answers such as “New Zealand.”

Improvement direction: redesign the query rewriting logic so that retrieval focuses on the country
where the game was first released, rather than the supported device platforms; ensure that the
final answer explicitly contains and highlights “New Zealand” as the only correct answer.
\end{Verbatim}
\end{tcolorbox}

However, the textual loss alone does not specify how an intermediate node should change. ADOPT therefore converts this final discrepancy into a \emph{global textual gradient}, which explains the failure mode of the end-to-end output at the workflow level.

\begin{tcolorbox}[breakable,colback=gray!10!white, colframe=gray!80!black,
  title=Global Textual Gradient, boxrule=0.5pt, arc=2mm, left=2mm, right=2mm, top=1mm, bottom=1mm]
\begin{Verbatim}[fontsize=\small, breaklines=true]
Analysis:
Although the external knowledge does not provide the specific factual content
(e.g., that the game Angry Birds was first released in New Zealand), we can still
analyze the root cause of the deviation between the current output and the gold answer
by combining question semantics, workflow behavior, and shallow discrepancy analysis.

1. The query rewriting stage failed to recognize that “Where” refers to a spatial attribute,
   causing retrieval to drift toward technical platforms.
   The system implicitly associated “launch” with “technical deployment” rather than a
   “geographic release event,” thereby guiding subsequent retrieval toward
   “which devices it ran on” rather than “which country it was first released in.”

2. The system lacked the ability to disentangle the temporal-spatial semantics of
   “first launch,” and thus confused different dimensions of release.
   It captured only the technical implementation aspect of “launch,” while failing to
   activate location-oriented reasoning implied by “first.” As a result, it did not produce
   retrieval queries explicitly targeting the geographic dimension.

3. The multi-step retrieval process failed to identify the information gap effectively,
   leading to missing geographic evidence.
   The final retrieved information contained no country or region name at all, yet the
   system did not trigger an additional location-oriented query.

4. The answer generation stage lacked an output validation mechanism aligned with
   the question type.
   The expected answer to a “Where” question should be a place name, whereas
   “iOS and Maemo devices” clearly does not match that pattern. However, the system
   failed to detect this structural inconsistency.

Overall, the root problem does not lie in a single node alone, but in the failure of the
entire workflow to consistently support geographic question answering across semantic
understanding, query formulation, information assessment, and output control. In particular,
the workflow failed to maintain and reinforce the spatial attribute implied by “Where,”
which caused it to drift toward an unintended technical interpretation path.
\end{Verbatim}
\end{tcolorbox}

The key step of ADOPT is then to combine the workflow-level diagnosis above with the dependency of \texttt{generate\_query} on the final answer, and derive a \emph{local textual gradient} for this specific non-final node. Unlike the global textual gradient, which characterizes the failure mode at the workflow-output level, the local textual gradient is step-specific, which attributes the relevant portion of the end-to-end error to \texttt{generate\_query} and specifies how the behavior of this node should change in order to improve the final answer. In other words, it transforms a final-output error signal into an optimization direction for an intermediate retrieval-oriented step.

\begin{tcolorbox}[breakable,colback=gray!10!white, colframe=gray!80!black,
  title=Local Textual Gradient for \texttt{generate\_query}, boxrule=0.5pt, arc=2mm, left=2mm, right=2mm, top=1mm, bottom=1mm]
\begin{Verbatim}[fontsize=\small, breaklines=true]
The past outputs of this node failed to guide the workflow toward the correct geographic answer, **“New Zealand,”** because the generated search queries were too broad and did not explicitly target the missing **country/location** dimension of the question.

In the first retrieval round, the query **“Where did the Angry Birds game first launch”** was grammatically valid, but it was ambiguous. It could easily be interpreted by the search system as asking about the **launch platform** (for example, iOS) rather than the **geographic location** of the first release. In the second round, **“Where did Angry Birds first launch when released in 2009?”** had the same problem: although it added a release year, it still did not constrain the search toward a **country**, **region**, or **launch location**, so the retrieval process remained misaligned.

To correct this node, the key change is to recognize how its previous outputs contributed to the final error. The node should identify that the workflow had already surfaced **technical release information** such as iOS or Maemo, but had still not resolved the **geographic aspect** of the question. Therefore, the missing information was not about platform or release date, but about **place**.

The parts of the past outputs that should be modified are the search queries themselves. Instead of using underspecified “Where” questions, the node should generate queries that explicitly encode the intended semantic dimension, such as **country**, **geographic location**, or **first release country**. This would better align retrieval with the actual answer requirement.

A better reasoning pattern for this node would be:

1. Recognize that the question is asking for a **geographic answer**.
2. Detect that existing context only provides **platform-level release details**, not **country-level information**.
3. Reformulate the query with explicit geographic constraints, such as:

   * **“Which country was Angry Birds first released in?”**
   * **“Angry Birds first released country 2009”**
   * **“Angry Birds initial launch location country”**

These modifications would make the node’s output more targeted and would increase the likelihood that later retrieval stages return the correct answer, **New Zealand**.
\end{Verbatim}
\end{tcolorbox}

However, the local textual gradient is still an abstract update direction expressed in natural language. To make it usable by a step-level prompt optimizer, ADOPT further converts this gradient into a \emph{revised step output}, i.e., a task-specific target output describing what \texttt{generate\_query} should have produced for this bad case. This revised output serves as pseudo-supervision for the intermediate node, even though no gold query is available in the original training data.

\begin{tcolorbox}[breakable,colback=gray!10!white, colframe=gray!80!black,
  title=Revised Step Output Derived from the Local Textual Gradient, boxrule=0.5pt, arc=2mm, left=2mm, right=2mm, top=1mm, bottom=1mm]
\begin{Verbatim}[fontsize=\small, breaklines=true]
Reasoning:
The output of the current node `generate_query` failed in both retrieval rounds to
effectively guide the system toward the correct geographic answer (“New Zealand”).
The root cause is that the generated `search_query` was too broad and did not explicitly
target the semantic dimension of “country” or “geographic location.”

The first query, “Where did the Angry Birds game first launch,” is grammatically correct,
but a search engine may interpret it as asking about release platforms (e.g., iOS)
rather than the place of first release. The second query,
“Where did Angry Birds first launch when released in 2009?”, remains ambiguous and
still fails to force retrieval results to focus on “country” or “location.”

According to the core responsibility of this node described in <DEPENDENCY>—namely,
to identify missing information and generate highly relevant, clearly directed retrieval
queries—the node should, given that the current context already reveals the technical
release platforms (iOS/Maemo), actively recognize that the key missing information is
still the geographic dimension, and then construct a more targeted search query.

Furthermore, based on the <MODIFICATION> analysis, the “Where” in the question should
be interpreted as a spatial location rather than a technical deployment environment.
Accordingly, the correct reasoning path should be:
1. Identify the question type as geographic (where → country / region);
2. Determine that the current context mentions only device platforms and lacks country information;
3. Construct a query containing geographic cues such as “country,” “first released in,”
   and “launch location” to redirect retrieval.

Therefore, the revised output should reflect the following improvements:
- The reasoning should clearly state that the current context provides release platform
  information but does not specify the country of first release;
- It should emphasize that a new query is needed to obtain the geographic answer;
- It should generate a precise `search_query`, for example using “which country” or
  “in what country,” to ensure that retrieval explicitly targets a country entity.

The final output should avoid simply repeating the original question and should instead
perform semantic strengthening and intent focusing, so as to push the workflow toward
the correct answer.

<REVISED_NODE_OUTPUT>
[[ ## reasoning ## ]]
The question asks for the geographic location where the Angry Birds game was first released,
rather than merely the platform on which it was launched. The current context states that the
game was first launched on iOS and Maemo devices in December 2009, which answers the
technical question of what kinds of devices it was released on, but does not specify the country
or region of first release. Since “launch” in this context should be understood as the geographic
starting point of market release rather than the runtime platform, further retrieval is needed to
determine in which country the game was originally released.

To fill this critical geographic information gap, the next query should explicitly focus on the
country-level answer, using phrases such as “which country” and “first released in” to guide
the retriever toward the relevant fact.

[[ ## search_query ## ]]
In which country was the Angry Birds game first released

[[ ## completed ## ]]
</REVISED_NODE_OUTPUT>
\end{Verbatim}
\end{tcolorbox}


The revised step output shown above serves as an illustrative example of how ADOPT transforms the local textual gradient for a single bad case into a target output for \texttt{generate\_query}. In the actual optimization process, this transformation is performed for multiple bad cases, yielding a collection of step-level input and output pairs that together form a step-specific dataset for the node. ADOPT then applies a single-prompt optimization procedure to this aggregated dataset to update the prompt of \texttt{generate\_query}.
The optimized prompt shown below is obtained from the full set of revised outputs collected across cases.

\begin{tcolorbox}[breakable,colback=gray!10!white, colframe=gray!80!black,
  title=Prompt of \texttt{generate\_query} After Optimization, boxrule=0.5pt, arc=2mm, left=2mm, right=2mm, top=1mm, bottom=1mm]
\begin{Verbatim}[fontsize=\small, breaklines=true]
Your input fields are:

1. `context` (str):  
A textual passage providing background information, potentially containing relevant entities,
dates, relationships, or contextual clues.

2. `question` (str):  
The user's information need, which may be direct, compound, ambiguous, or contain misstated concepts.

Your output fields are:

1. `reasoning` (str):  
A structured explanation that MUST include:
- Semantic analysis of the question, identifying its type
- Decomposition of complex questions into sequential sub-questions
- Diagnosis of information gaps relative to the provided context
- Justification for query design choices
- Fallback strategy if the context is insufficient

2. `search_query` (str):  
A concise, search-engine-optimized query string that:
- Uses minimal, high-signal keywords
- Avoids full sentences or conversational phrasing
- Applies precise terminology based on semantic clarification
- Incorporates exact matches when appropriate
- Targets one primary retrieval intent

All interactions will follow this strict format:

[[ ## context ## ]]
{context}

[[ ## question ## ]]
{question}

[[ ## reasoning ## ]]
{reasoning}

[[ ## search_query ## ]]
{search_query}

[[ ## completed ## ]]

Objective:
Given `context` and `question`, produce `reasoning` and `search_query` such that the latter
enables efficient retrieval of the answer through standard search engines.

You must not assume external tools are available; your sole function is to generate an optimal
search query grounded in rigorous, transparent reasoning.
\end{Verbatim}
\end{tcolorbox}


This example highlights the core advantage of ADOPT. Even though the system does not observe any intermediate supervision for \texttt{generate\_query}, it can still optimize this non-final LLM step by: (1) identifying, through dependency analysis, how this node contributes to end-to-end success; (2) converting the final-answer error into a global textual gradient that captures the workflow-level failure mode; (3) decomposing that signal into a local textual gradient; and (4) converting this local gradient into a revised step output that serves as pseudo-supervision for prompt updating. As a result, optimization is not restricted to the final answer module, but can effectively improve upstream retrieval-oriented steps whose outputs only influence the task indirectly through later pipeline execution.


\newpage
\onecolumn

\section{Implementation Details}
\label{sec:appendix}
Specifically, as shown in Table~\ref{tab:table-data}, each multi-step pipelines is implemented as follows:

\vspace{0.05in}
\noindent\textbf{Multi-step Pipeline for Amazon.}
For Amazon, we use 3 LLM steps in the pipeline. Two LLM steps are used for summarizing information from an item and a user, and the other LLM step is used to generate the possible rating based on the summarization of the item and user.

\begin{tcolorbox}[breakable,colback=gray!10!white, colframe=gray!80!black,
  title=Pseudocode of Pipeline for Amazon, boxrule=0.5pt, arc=2mm, left=2mm, right=2mm, top=1mm, bottom=1mm]
\begin{Verbatim}[fontsize=\small, commandchars=\\\{\}]
Initialize Module:
    initialize LLM step \textcolor{blue}{conclude_item}
    initialize LLM step \textcolor{blue}{conclude_user}
    initialize LLM step \textcolor{blue}{rating}

Function forward(inputs):
    # Step 1: derive item-centric conclusion from item reviews
    conclusion_1 ← \textcolor{blue}{conclude_item}(
                        product = inputs.product,
                        reviews_item = inputs.reviews_item
                    )

    # Step 2: derive user-centric conclusion from user reviews
    conclusion_2 ← \textcolor{blue}{conclude_user}(
                        reviews_user = inputs.reviews_user
                    )

    # Step 3: aggregate conclusions and produce final rating
    stars ← \textcolor{blue}{rating}(
                 user = inputs.user,
                 product = inputs.product,
                 conclusion_1 = conclusion_1,
                 conclusion_2 = conclusion_2
             )

    # Output
    return stars
\end{Verbatim}
\end{tcolorbox}

\vspace{0.05in}
\noindent\textbf{Multi-step Pipeline for HotPotQA.}
We use a two-step LLM pipeline for HotPotQA. The generate\_query module iteratively generates retrieval queries based on the question and current context for multi-hop retrieval, while the generate\_answer module produces the final answer using the question and all retrieved context.

\begin{tcolorbox}[breakable,colback=gray!10!white, colframe=gray!80!black,
  title=Pseudocode of Pipeline for HotPotQA, boxrule=0.5pt, arc=2mm, left=2mm, right=2mm, top=1mm, bottom=1mm]
\begin{Verbatim}[fontsize=\small, commandchars=\\\{\}]
Initialize Agent:
    initialize LLM step \textcolor{blue}{generate_query}
    initialize LLM step \textcolor{blue}{generate_answer}
    initialize retriever with top_k = 3

Function forward(inputs):
    context ← empty list

    repeat for N hops (N = 2):
        query ← \textcolor{blue}{generate_query}(
                    question = inputs.question,
                    context = context
                )
                
        processed_query ← process(query)
        retrieved_messages ← retrieve(processed_query)
        context.append(retrieved_messages)

    answer ← \textcolor{blue}{generate_answer}(
                 question = inputs.question,
                 context = context
             )

    return answer
\end{Verbatim}
\end{tcolorbox}

\vspace{0.05in}
\noindent\textbf{Multi-step Pipeline for HoVer.}
For HoVer, we use a multi-step LLM pipeline with explicit stages. The pipeline first summarizes initial retrieved evidence using summarize\_hop1. Based on this summary, create\_query\_hop2 generates a refined query for second-hop retrieval, whose results are further condensed by summarize\_hop2. Next, create\_query\_hop3 produces an expanded query to retrieve additional evidence. Finally, the judge module verifies the claim by reasoning over all retrieved documents.

\begin{tcolorbox}[breakable,colback=gray!10!white, colframe=gray!80!black,
  title=Pseudocode of Pipeline for HoVer, boxrule=0.5pt, arc=2mm, left=2mm, right=2mm, top=1mm, bottom=1mm]
\begin{Verbatim}[fontsize=\small, commandchars=\\\{\}]
Initialize Agent:
    initialize LLM step \textcolor{blue}{create_query_hop2}
    initialize LLM step \textcolor{blue}{create_query_hop3}
    initialize LLM step \textcolor{blue}{summarize_hop1}
    initialize LLM step \textcolor{blue}{summarize_hop2}
    initialize LLM step \textcolor{blue}{judge}
    initialize retriever with top_k = 5

Function forward(inputs):
    claim ← inputs.claim

    # Stage 1: initial evidence retrieval and summarization
    hop1_docs ← retrieve(claim)

    summary_1 ← \textcolor{blue}{summarize_hop1}(
                    claim = claim,
                    passages = hop1_docs
                )

    # Stage 2: summary-conditioned retrieval and refinement
    hop2_query ← \textcolor{blue}{create_query_hop2}(
                     claim = claim,
                     summary_1 = summary_1
                 )

    hop2_docs ← retrieve(hop2_query)

    summary_2 ← \textcolor{blue}{summarize_hop2}(
                    claim = claim,
                    context = summary_1,
                    passages = hop2_docs
                )

    # Stage 3: iterative query expansion and evidence retrieval
    hop3_query ← \textcolor{blue}{create_query_hop3}(
                     claim = claim,
                     summary_1 = summary_1,
                     summary_2 = summary_2
                 )

    hop3_docs ← retrieve(hop3_query)

    # Stage 4: evidence-based claim verification
    all_docs ← hop1_docs + hop2_docs + hop3_docs

    result ← \textcolor{blue}{judge}(
                 claim = claim,
                 context = all_docs
             )

    # Output
    return result
\end{Verbatim}
\end{tcolorbox}

\vspace{0.05in}
\noindent\textbf{Multi-step Pipeline for Emergency.}
For Emergency, we design an iterative dialogue-based pipeline with four explicit LLM components. The chatbot module generates clinically relevant questions based on the dialogue history and current patient information to interact with a simulated patient. Patient responses are processed by extract\_patient\_info to extract and update structured clinical information. The check\_info\_sufficiency module continuously evaluates whether the collected information is sufficient to make a triage decision and identifies missing required information to guide further questioning. Once sufficient information is obtained, the triage\_classifier assigns the final ATS category based on the extracted patient information.

\begin{tcolorbox}[breakable,colback=gray!10!white, colframe=gray!80!black,
  title=Pseudocode of Pipeline for Emergency, boxrule=0.5pt, arc=2mm, left=2mm, right=2mm, top=1mm, bottom=1mm]
\begin{Verbatim}[fontsize=\small, commandchars=\\\{\}]
Initialize Module:
    initialize LLM step \textcolor{blue}{chatbot}
    initialize LLM step \textcolor{blue}{extract_patient_info}
    initialize LLM step \textcolor{blue}{check_info_sufficiency}
    initialize LLM step \textcolor{blue}{triage_classifier}

Initialize Patient:
    # Simulate a human patient based on the information in the dataset
    initialize empty dialogue history

Function forward(inputs):
    # Initialize simulated patient with a given clinical scenario
    patient ← Patient(inputs.scenario)

    # Initialize structured patient information schema
    patient_info ← \{
        name: "",
        age: "",
        gender: "",
        presenting_problem: "",
        associated_symptoms: "",
        primary_survey: \{
            Airway: "",
            Breathing: "",
            Circulation: "",
            Disability: "",
            Environment: ""
        \},
        focused_assessment: "",
        pertinent_history: "",
        red_flags: ""
    \}

    # Initialize dialogue control variables
    round_count ← 1
    max_rounds ← 15
    conversation_history ← empty list
    next_question ← ""
    required_info ← ""
    all_required_info ← empty list

    # Iterative information-seeking dialogue
    while round_count <= max_rounds:

        # Step 1: obtain patient response
        if round_count = 1:
            answer ← initial patient utterance
        else:
            answer ← patient.get_answer(next_question)

        conversation_history.append("Patient: " + answer)

        # Step 2: extract structured clinical information from dialogue
        if round_count > 1:
            extracted_info ← \textcolor{blue}{extract_patient_info}(
                                 patient_info = patient_info,
                                 dialogue = conversation_history
                             )

            if extracted_info is valid:
                patient_info ← parsed extracted_info

        # Step 3: generate next clinical question
        next_question ← \textcolor{blue}{chatbot}(
                            patient_info = patient_info,
                            dialogue = conversation_history,
                            required_info = required_info
                        )

        conversation_history.append("Clinical Staff: " + next_question)

        round_count ← round_count + 1

        # Step 4: assess information sufficiency for triage decision
        is_sufficient, required_info ← \textcolor{blue}{check_info_sufficiency}(
                                            previously_requested = all_required_info,
                                            patient_info = patient_info
                                        )

        if is_sufficient:
            break

        all_required_info.append(required_info)

    # Final Step: assign triage category based on collected patient information
    ATS_category ← \textcolor{blue}{triage_classifier}(
                       patient_info = patient_info
                   )

    return ATS_category
\end{Verbatim}
\end{tcolorbox}

\vspace{0.05in}
\noindent\textbf{Multi-step Pipeline for Harmful.}
For harmful meme detection, we adopt a multi-step discussion-based pipeline with four explicit LLM components. The internet\_user and internet\_supervisor modules iteratively discuss and query different aspects of the meme based on structured image descriptions, jointly exploring potentially harmful content. Their interaction is coordinated by check\_info\_sufficiency, which determines when sufficient information has been gathered to terminate the discussion. Finally, the summary module analyzes the complete discussion history and aggregated meme information to produce the final harmfulness assessment.
All image-related information is obtained by calling an independent VQA module based on Qwen3-VL-Plus through the describe\_image, extract\_text, recognize\_celebrities, and get\_answer\_from\_image functions.

\begin{tcolorbox}[breakable,colback=gray!10!white, colframe=gray!80!black,
  title=Pseudocode of Pipeline for Harmful, boxrule=0.5pt, arc=2mm, left=2mm, right=2mm, top=1mm, bottom=1mm]
\begin{Verbatim}[fontsize=\small, commandchars=\\\{\}]
Initialize Workflow:
    initialize LLM step \textcolor{blue}{internet_user}
    initialize LLM step \textcolor{blue}{internet_supervisor}
    initialize LLM step \textcolor{blue}{check_info_sufficiency}
    initialize LLM step \textcolor{blue}{summary}

Function forward(inputs):
    # Extract multimodal meme information
    meme_description ← describe_image(inputs.image_path)
    meme_text ← extract_text(inputs.image_path)
    celebrity_info ← recognize_celebrities(inputs.image_path)

    # Assemble structured meme description
    description ← (meme_description, meme_text, celebrity_info)

    # Initialize dialogue history
    history ← empty dictionary

    # Define participating LLM steps
    characters ← ["Internet User", "Internet Supervisor"]
    LLM_steps ← \{
        "Internet User": \textcolor{blue}{internet_user},
        "Internet Supervisor": \textcolor{blue}{internet_supervisor}
    \}
    round ← 1

    # Multi-round, multi-characters discussion loop
    while round <= 10:
        for each character in characters:
            # Construct dialogue context from history
            dialogue_context ← format(history)

            # Generate character-specific question
            question ← LLM_steps[character](
                            description = description,
                            message = dialogue_context
                        )

            # Answer the question using image-grounded VQA
            answer ← get_answer_from_image(image_path, query = question)

            # Record question–answer pair
            history["Round " + round] ←
                "Question (" + character + "): " + question + " | Answer: " + answer

        round ← round + 1

        # Check whether collected information is sufficient
        is_sufficient ← \textcolor{blue}{check_info_sufficiency}(
                            history = history
                        )

        if is_sufficient:
            break

    # Generate final summary based on meme information and QA history
    result ← \textcolor{blue}{summary}(
                 description = description,
                 qa_history = history
             )

    return result
\end{Verbatim}
\end{tcolorbox}

\newpage
\onecolumn

\section{Prompts of Optimizers}
\label{sec:appendix-prompt}

\begin{tcolorbox}[breakable,colback=gray!10!white, colframe=gray!80!black,
  title=System Prompt of optimizer $E_1$, boxrule=0.5pt, arc=2mm, left=2mm, right=2mm, top=1mm, bottom=1mm]
\begin{Verbatim}[fontsize=\small, breaklines=true]
## Role
You are a master of LLM-workflow analysis, capable of precisely identifying the task of the entire workflow, the responsibility of every LLM node, and the dependency between each node’s output and the final result.

## Information Provided
The complete forward-pass code of the LLM workflow, if there is no forward-pass code, skip it.
The prompt of each LLM node (members of self.nodes)

## Skills
Deduce the overall task the LLM workflow is solving
Determine the specific duty of each LLM node
Clarify how the quality of each node’s output influences the end-to-end final result

## Deliverables
The task the workflow accomplishes
The responsibility of each LLM node
The correlation and impact of each node on the final output
\end{Verbatim}
\end{tcolorbox}

\begin{tcolorbox}[breakable,colback=gray!10!white, colframe=gray!80!black,
  title=User Prompt of optimizer $E_1$, boxrule=0.5pt, arc=2mm, left=2mm, right=2mm, top=1mm, bottom=1mm]
\begin{Verbatim}[fontsize=\small, breaklines=true]
## forward-pass code of the LLM workflow
```python
{{forward_code}}
```

## system prompt of current node
{{system_prompt}}}}

## user prompt of current node
{{user_prompt}}}}
\end{Verbatim}
\end{tcolorbox}

\begin{tcolorbox}[breakable,colback=gray!10!white, colframe=gray!80!black,
  title=System Prompt of optimizer $E_2$, boxrule=0.5pt, arc=2mm, left=2mm, right=2mm, top=1mm, bottom=1mm]
\begin{Verbatim}[fontsize=\small, breaklines=true]
## Role  
You are a workflow-analysis master who excels at pinpointing how a **single LLM call** affects the **final workflow output**.  
Based on the **inputs & outputs of that LLM node** and the **final result** in **multiple good cases**, as well as a rough summary of the LLM node's duties, you further refine the **exact responsibility** of that LLM call (a member of `self.nodes`).

## Information Received  
You will receive:  
1. The name of the LLM node to be analyzed (`self.nodes[xxxxx]`)  
2. A rough summary of the LLM workflow and its nodes  
3. The prompt of the LLM node to be analyzed  
4. Multiple good cases in JSON format:  
   - input to that LLM call  
   - output of that LLM call  
   - workflow input  
   - workflow output  
   - ground-truth answer  
   - result judgment  

## Skills  
1. Carefully analyze **every good case** provided and summarize commonalities.  
2. Combine the case data to determine **what specific duty** the LLM node performs when the workflow runs correctly, and **what contribution** its output makes to the end-to-end result.  
3. Combine the case data to reason about **how the LLM node’s output correlates with the final output**.  
4. Consider how **changes in the LLM node’s output** might **positively or negatively impact** the final workflow result.

## Notes  
When analyzing duties, contributions, and impacts, **synthesize across all good cases** to produce **generalizable** conclusions.

## Answer Content  
You must provide a detailed answer to:  
1. Based on the good cases, what **specific duty** does this LLM node perform in the workflow? What contribution does its output make to the final end-to-end result?  
2. Based on the good cases, what is the **correlation** between this LLM node’s output and the final output?  
3. Based on the good cases, how might **changes in the LLM node’s output** positively or negatively affect the final workflow output?

## Output
The summary of node responsibility should be concise.
\end{Verbatim}
\end{tcolorbox}

\begin{tcolorbox}[breakable,colback=gray!10!white, colframe=gray!80!black,
  title=User Prompt of optimizer $E_2$, boxrule=0.5pt, arc=2mm, left=2mm, right=2mm, top=1mm, bottom=1mm]
\begin{Verbatim}[fontsize=\small, breaklines=true]
### Current LLM call to be summarized:
{{node_name}}

### Rough summary of the LLM agent/workflow
{{agent_description}}

### System prompt of the current LLM call:
{{system_prompt}}

### User prompt of the current LLM call:
{{user_prompt}}

### In various good cases, the input & output of this LLM call, the workflow input, the workflow output, the ground-truth answer, and the result judgment (in JSON format):
{{good_cases}}
\end{Verbatim}
\end{tcolorbox}

\begin{tcolorbox}[breakable,colback=gray!10!white, colframe=gray!80!black,
  title=System Prompt of optimizer $E_3$, boxrule=0.5pt, arc=2mm, left=2mm, right=2mm, top=1mm, bottom=1mm]
\begin{Verbatim}[fontsize=\small, breaklines=true]
You are the dedicated feedback engine for output of a multi-stage workflow.
Your only responsibility is to analyze a single candidate response and produce a constructive, metric-driven feedback that, when applied, maximizes its score under a single metric.
You don't need to consider optimizing any node; you only need to focus on modifying the output.

Specifications:
1. The workflow description is provided in <WORKFLOW_DESCRIPTION> {{workflow_description}} </WORKFLOW_DESCRIPTION>.
2. The current workflow output is provided in <CURRENT_OUTPUT> {{current_output}} </CURRENT_OUTPUT>.
3. The expected correct output is provided in <GROUND_TRUTH> {{ground_truth}} </GROUND_TRUTH>.
4. The evaluation function is <METRIC_FUNCTION> {{metric_fn}} </METRIC_FUNCTION>.
5. You should only critique the candidate, not rewrite it. Focus on actionable suggestions.
6. List 1-5 specific differences between the candidate response and the ground truth that are causing the score to drop, quoting the exact tokens or fields. If the metric function checks for exact or approximate equality, you **must** list **every** specific difference between output and the ground truth, quoting each differing field or token.
7. If the candidate already achieves a perfect score (1.0), reply: "No improvement needed, optimal score achieved".
8. Limit feedback to actionable suggestions for improving the specified metric.
9. **important** Strictly adhere to the given optimization <CONSTRAIN>.
\end{Verbatim}
\end{tcolorbox}

\begin{tcolorbox}[breakable,colback=gray!10!white, colframe=gray!80!black,
  title=User Prompt of optimizer $E_3$, boxrule=0.5pt, arc=2mm, left=2mm, right=2mm, top=1mm, bottom=1mm]
\begin{Verbatim}[fontsize=\small, breaklines=true]
Here is the information for your feedback task:

- <WORKFLOW_DESCRIPTION> {{workflow_description}} </WORKFLOW_DESCRIPTION>
- <CURRENT_OUTPUT> {{current_output}} </CURRENT_OUTPUT>
- <GROUND_TRUTH> {{ground_truth}} </GROUND_TRUTH>
- <METRIC_FUNCTION> {{metric_fn}} </METRIC_FUNCTION>
- <CURRENT_SCORE> {{current_score}} </CURRENT_SCORE>
- <CONSTRAIN> {{constrain}} </CONSTRAIN>

<OBJECT>
1. Provide actionable, metric-focused feedback to increase the candidate response's score.
2. Identify 1-5 key differences between the candidate response and the ground truth. quoting the relevant tokens or fields.
</OBJECT>
\end{Verbatim}
\end{tcolorbox}

\begin{tcolorbox}[breakable,colback=gray!10!white, colframe=gray!80!black,
  title=System Prompt of optimizer $E_4$, boxrule=0.5pt, arc=2mm, left=2mm, right=2mm, top=1mm, bottom=1mm]
\begin{Verbatim}[fontsize=\small, breaklines=true]
You are the deep-dive analysis assistant for workflow outputs.
Your task is to explain **why** the workflow's actual output deviates from the expected output, using only the provided external knowledge-without considering or replying on nodes' prompt.

Requirements:
1. Begin with an "Analysis:" section where you think step by step, citing specific fragments from the external knowledge and showing how they apply (or were violated) in the actual output.
2. Then list reasons for each different.
3. Each reason must:
    - Quote the relevant excerpt from <EXTERNAL_KNOWLEDGE>.
    - Point out the discrepancy between the excerpt and the workflow's <CURRENT_OUTPUT> versus the <GROUND_TRUTH>.
    - Explain how that gap led to the incorrect output.
4. Do **not** propose any fixes or mention the prompt-only diagnose the failure of the output against the external knowledge.
5. **important** Strictly adhere to the given optimization <CONSTRAIN>.
\end{Verbatim}
\end{tcolorbox}

\begin{tcolorbox}[breakable,colback=gray!10!white, colframe=gray!80!black,
  title=User Prompt of optimizer $E_4$, boxrule=0.5pt, arc=2mm, left=2mm, right=2mm, top=1mm, bottom=1mm]
\begin{Verbatim}[fontsize=\small, breaklines=true]
Here is the information for your analysis:

- <WORKFLOW_DESCRIPTION> {{workflow_description}} </WORKFLOW_DESCRIPTION>
- <INPUT> {{node_input}} </INPUT>
- <CURRENT_OUTPUT> {{node_output}} </CURRENT_OUTPUT>
- <GROUND_TRUTH> {{node_expected_output}} </GROUND_TRUTH>
- <EXTERNAL_KNOWLEDGE>
- <CONSTRAIN> {{constrain}} </CONSTRAIN>
{{external_knowledge}}
</EXTERNAL_KNOWLEDGE>
- <SHALLOW_DIFFERENCE>
{{shallow_difference}}
</SHALLOW_DIFFERENCE>

<OBJECT>
Provide deep reasons why the <CURRENT_OUTPUT> fails to comply with <EXTERNAL_KNOWLEDGE>, building on-nut not repeating-the surface mismatches listed in <SHALLOW_DIFFERENCE>, thereby explaining the root causes of the discrepancy with <GROUND_TRUTH>.
</OBJECT>
\end{Verbatim}
\end{tcolorbox}

\begin{tcolorbox}[breakable,colback=gray!10!white, colframe=gray!80!black,
  title=System Prompt of optimizer $E_5$, boxrule=0.5pt, arc=2mm, left=2mm, right=2mm, top=1mm, bottom=1mm]
\begin{Verbatim}[fontsize=\small, breaklines=true]
You are the optimization assistant for a specific LLM node within a multi-stage workflow.

Requirements:
You need to consider how the past outputs of this node led to the final erroneous result, then think about which parts of the past outputs should be modified to correct the final erroneous result.
\end{Verbatim}
\end{tcolorbox}

\begin{tcolorbox}[breakable,colback=gray!10!white, colframe=gray!80!black,
  title=User Prompt of optimizer $E_5$, boxrule=0.5pt, arc=2mm, left=2mm, right=2mm, top=1mm, bottom=1mm]
\begin{Verbatim}[fontsize=\small, breaklines=true]
Here is the information for your task:

- <DEPENDENCY> {{dependency_from_this_workflow_final_output}} </DEPENDENCY> (Dependency and job description, how this node's output affects the final output)
- <CURRENT_WORKFLOW_OUTPUT> {{workflow_output}} </CURRENT_WORKFLOW_OUTPUT> (Output of the entire workflow, the output is in JSON format.)
- <MODIFICATION> {{modification}} </MODIFICATION> (The **modification** for the workflow output and **why** it is wrong)
- <NODE_IN_BLOCK> {{node_in_block}} </NODE_IN_BLOCK> (The input is in JSON format, and output is in STRING format.)

<OBJECT>
Think step by step and then produce all of your thinking.
</OBJECT>
\end{Verbatim}
\end{tcolorbox}

\begin{tcolorbox}[breakable,colback=gray!10!white, colframe=gray!80!black,
  title=System Prompt of optimizer $E_6$, boxrule=0.5pt, arc=2mm, left=2mm, right=2mm, top=1mm, bottom=1mm]
\begin{Verbatim}[fontsize=\small, breaklines=true]
You are the optimization assistant for a specific LLM node within a multi-stage workflow.

Requirements:
1. Begin with your chain of thought under the heading "Reasoning:".
2. After your reasoning, produce the final node output wrapped exactly in <REVISED_NODE_OUTPUT>...</REVISED_NODE_OUTPUT> tags.
3. Do not include any other text outside the "Reasoning:" section and the tagged output.
4. Do **not** write the REVISED_NODE_OUTPUT in the same way a CURRENT_WRONG_NODE_OUTPUT.
5. You need to first consider how the past outputs of this node led to the final erroneous result, then think about which parts of the past outputs should be modified to correct the final erroneous result, and finally write the revised output.
\end{Verbatim}
\end{tcolorbox}

\begin{tcolorbox}[breakable,colback=gray!10!white, colframe=gray!80!black,
  title=User Prompt of optimizer $E_6$, boxrule=0.5pt, arc=2mm, left=2mm, right=2mm, top=1mm, bottom=1mm]
\begin{Verbatim}[fontsize=\small, breaklines=true]
Here is the information for your task:

- <DEPENDENCY> {{dependency_from_this_workflow_final_output}} </DEPENDENCY> (Dependency and job description, how this node's output affects the final output)
- <CURRENT_WORKFLOW_OUTPUT> {{workflow_output}} </CURRENT_WORKFLOW_OUTPUT> (Output of the entire workflow, the output is in JSON format.)
- <MODIFICATION> {{modification}} </MODIFICATION> (The **modification** for the workflow output and **why** it is wrong)
- <NODE_IN_BLOCK> {{node_in_block}} </NODE_IN_BLOCK> (The input is in JSON format, and output is in STRING format.)
- <Reasoning> {{thoughts}} </Reasoning> (You can keep thinking based on your past thoughts.)
- <REVISED_NODE_OUTPUT>...</REVISED_NODE_OUTPUT> (The output is in STRING format. You should give revised output difference from CURRENT_WRONG_NODE_OUTPUT to make the **modification** for the workflow output based on the DEPENDENCY)

<OBJECT>
Think step by step ("Reasoning:") and then produce all the exact text this node should emit, wrapped in <REVISED_NODE_OUTPUT>...</REVISED_NODE_OUTPUT> so it can be programmatically extracted.
</OBJECT>
\end{Verbatim}
\end{tcolorbox}

\end{document}